\documentclass{article}

\usepackage{microtype}
\usepackage{graphicx}
\usepackage{subfigure}
\usepackage{booktabs} %
\usepackage{amsmath}
\usepackage{makecell}
\usepackage{multicol}
\usepackage{multirow}
\usepackage[english]{babel}

\usepackage{amsmath,amsfonts,bm}

\def\eqref#1{equation~\ref{#1}}

\def\ceil#1{\lceil #1 \rceil}
\def\floor#1{\lfloor #1 \rfloor}
\def\1{\bm{1}}

\DeclareMathAlphabet{\mathsfit}{\encodingdefault}{\sfdefault}{m}{sl}
\SetMathAlphabet{\mathsfit}{bold}{\encodingdefault}{\sfdefault}{bx}{n}

\def\emW{{W}}

\newcommand{\softmax}{\mathrm{softmax}}
\newcommand{\sigmoid}{\sigma}

\def\talkfull{Time-aware Large Kernel Convolution}

\def\talkfullsabbr{Time-aware Large Kernel (TaLK) Convolutions}
\def\talkconvfull{TaLK Convolution}
\def\talkconvfulls{TaLK Convolutions}

\def\wiki{WikiText-103}
\def\cnndm{CNN-DailyMail}
\def\iwslt{IWSLT De-En}
\def\ende{WMT En-De}
\def\enfr{WMT En-Fr}

\usepackage{hyperref}

\usepackage[accepted]{icml2020}

\icmltitlerunning{Time-aware Large Kernel Convolutions}

\begin{document}

\twocolumn[
\icmltitle{Time-aware Large Kernel Convolutions}

\icmlsetsymbol{equal}{*}

\begin{icmlauthorlist}
\icmlauthor{Vasileios Lioutas}{to}
\icmlauthor{Yuhong Guo}{to}
\end{icmlauthorlist}

\icmlaffiliation{to}{School of Computer Science, Carleton University, Canada}

\icmlcorrespondingauthor{Vasileios Lioutas}{contact@vlioutas.com}

\icmlkeywords{Sequence Modeling, Attention, Convolutional Sequence Modeling}

\vskip 0.3in
]

\printAffiliationsAndNotice{}  

\begin{abstract}
To date, most state-of-the-art sequence modeling architectures use attention to build generative models for language based tasks. Some of these models use all the available sequence tokens to generate an attention distribution which results in time complexity of $O(n^2)$. Alternatively, they utilize depthwise convolutions with $\softmax$ normalized kernels of size $k$ acting as a limited-window self-attention, resulting in time complexity of $O(k{\cdot}n)$. In this paper, we introduce Time-aware Large Kernel (TaLK) Convolutions, a novel adaptive convolution operation that learns to predict the size of a summation kernel instead of using a fixed-sized kernel matrix. This method yields a time complexity of $O(n)$, effectively making the sequence encoding process linear to the number of tokens. We evaluate the proposed method on large-scale standard machine translation, abstractive summarization and language modeling datasets and show that TaLK Convolutions constitute an efficient improvement over other attention/convolution based approaches.
\end{abstract}

\section{Introduction}

Sequence modeling has seen some great breakthroughs through recent years with the introduction of the use of neural networks. Recurrent neural network methods \cite{DBLP:journals/corr/SutskeverVL14,bahdanau2014neural,DBLP:journals/corr/WuSCLNMKCGMKSJL16}, convolution methods \cite{kim-2014-convolutional,kalchbrenner-etal-2014-convolutional,DBLP:journals/corr/KalchbrennerESO16,gehring2017convolutional,wu2019pay}, and self-attention approaches \cite{DBLP:journals/corr/PaulusXS17,vaswani2017attention,DBLP:journals/corr/abs-1901-02860, kitaev2020reformer} have all yielded state-of-the-art results in various NLP tasks such as neural machine translation (NMT) \cite{DBLP:journals/corr/SutskeverVL14,DBLP:journals/corr/WuSCLNMKCGMKSJL16,DBLP:journals/corr/BritzGLL17,DBLP:journals/corr/abs-1903-00089}, language modeling \cite{Sundermeyer2012,DBLP:journals/corr/TranBM16,DBLP:journals/corr/abs-1810-04805,radford2019language}, automatic summarization \cite{DBLP:journals/corr/PaulusXS17,DBLP:journals/corr/abs-1711-05217,DBLP:journals/corr/abs-1803-10357}, named entity recognition \cite{DBLP:journals/corr/LampleBSKD16,DBLP:journals/corr/abs-1810-04805} and  sentiment analysis \cite{DBLP:journals/corr/XuCQH16,Sachan2019RevisitingLN}.

Seemingly all modern approaches of sequence encoding rely on the use of attention to ``filter'' the excessive information given at a current time-step. Attention can be expressed as the weighted sum over context representations using attention weights that are usually generated from the context representations (self-attention) \cite{Cheng_2016}. The transformer network \cite{vaswani2017attention} assigns attention weights for a given time-step to all available context token representations, while the newly proposed dynamic convolution \cite{wu2019pay} only computes an attention over a fixed context window. 

Self-attention over all context tokens is computationally very expensive. Specifically, the transformer network has a time complexity of $O(n^2)$ where $n$ is the length of the input sequence. Thus, modeling long-range dependencies becomes very challenging and the practicality of the self-attention method has been questioned.
The more recent approach of dynamic convolutions \cite{wu2019pay} successfully reduced the time complexity to $O(k{\cdot}n)$ where $k$ is 
the kernel size specified for each layer.

In this paper, we introduce a novel type of adaptive convolution, Time-aware Large Kernel (TaLK) convolutions, that learns the kernel size of a summation kernel for each time-step instead of learning the kernel weights as in a typical convolution operation. For each time-step, a function is responsible for predicting the appropriate size of neighbor representations to use in the form of left and right offsets relative to the time-step. The result is an efficient encoding method that reduces the time complexity to $O(n)$ and uses fewer parameters than all other methods. The method employs the fast Parallel Prefix Sum \cite{10.1145/322217.322232,18021} operation which has a time complexity of $O(\log(n))$
to compute the \textit{integral image} \cite{lewis1994}, also known as \textit{summed-area table} in the Computer Vision literature. This needs to be computed only once and can be used to calculate any summation between two boundary tokens in $O(1)$. 
Applying it on a sequence with length $n$ only needs $O(n)$ time. 
To summarize, the contributions of this work are three-fold:
\vskip -.3in
\begin{itemize}
\item   
We introduce a novel adaptive convolution based on summation kernel for sequence encoding. 
\vskip -.3in
\item   
We show both analytically and empirically that 
the proposed kernel method has a smaller time complexity;
it is faster than previous state-of-the-art approaches and is able to encode longer sentences quicker and with a smaller running memory footprint.
\vskip -.3in
\item   
We evaluate our method on three NLP tasks, machine translation, abstractive summarization and language modeling. We show that the proposed method can get comparative performance with previous methods on \ende{} and \enfr{} benchmarks in machine translation, and set a new state-of-the-art result on the \iwslt{} and \cnndm{} datasets, while in language modeling our method is able to perform comparatively with self-attention and outperform dynamic convolutions on the \wiki{} benchmark dataset.
\end{itemize}
Our code and pre-trained models are available at \href{https://github.com/lioutasb/TaLKConvolutions}{github.com/lioutasb/TaLKConvolutions}.

\section{Related Work}

In this section, we provide a brief review over various related sequence modeling methods, and related methods that enlarge the receptive filed of a convolution operation.

\subsection{Sequence Modeling}
Sequence modeling is an important task in machine learning. 
An effective
system should be able to comprehend and generate sequences similar to real data. Traditional approaches typically rely on the use of various kinds of recurrent neural networks such as long-short term memory networks \cite{doi:10.1162/neco.1997.9.8.1735,DBLP:journals/corr/SutskeverVL14,li-etal-2016-persona,LI2018301} and gated recurrent unit networks \cite{DBLP:journals/corr/ChoMGBSB14,nabil-etal-2016-cufe}. These recurrent approaches are auto-regressive, which slows the process down for long sequences since they linearly depend on their own previous output tokens. Recent work is focused on exploring convolutional neural networks (CNN) methods \cite{DBLP:journals/corr/KalchbrennerESO16,gehring2017convolutional,wu2019pay} or self-attention methods \cite{vaswani2017attention,zhang-Etal:2018:ACL2018accelerating,DBLP:journals/corr/abs-1901-02860,kitaev2020reformer} which both facilitate the parallelization of the encoding process. 
In addition, since they are not auto-regressive, they allow the encoding process to capture stronger global and local dependencies.

Recently, \citet{wu2019pay} proposed an alternative method to the original self-attention approach. Their window-based attention method with window size $k$ can perform comparatively with self-attention modules that have access to all available tokens at each time-step. They utilize a depthwise convolution with a generated $\softmax$ normalized kernel of size $k$ for every time-step. This brings down the time complexity to $O(k{\cdot}n)$ from the quadratic complexion of the original self-attention model. However, this method has the drawback of being memory intensive for long sequences depending on the implementation used. Moreover, supporting larger kernel sizes can have a negative impact to running time. In another work, \citet{DBLP:journals/corr/abs-1804-00857} proposed computing intra-block self-attention weights within blocks of the input sequence and inter-block attention between all blocks to reduce the running memory of the full self-attention approach.

Our method differs from all these previous approaches in two main aspects. Specifically, instead of having all (or some) tokens available and then deciding which ones are needed to encode a time-step, we start from the current time-step representation and try to expand to the neighbor tokens in an adaptive manner.  Additionally, instead of using attention for filtering the tokens used for encoding a time-step, we use all the information available in an adaptively decided window by utilizing a summation convolution kernel with summed-area tables, which improves upon previously proposed methodology by allowing us to reduce the complexity to $O(n)$, to produce a smaller running memory footprint, and to use less parameters than all other methods.

\subsection{Dynamically Sized Receptive Field}
Increasing the receptive field of a convolution layer without adding a computation overhead is a challenging task. By making deeper CNN models, we may be able to accumulate many fixed-sized receptive fields, however this comes at the cost of high computational demands. Nevertheless, this approach is shown to be successful in multiple state-of-the-art vision models \cite{DBLP:journals/corr/HeZRS15,DBLP:journals/corr/SzegedyIV16}. The overhead issue is often mitigated using a form of downsampling, either via pooling layers \cite{726791} or strided convolutions \cite{springenberg2014striving}. \citet{yu2015multiscale} proposed dilated convolutions, a method for enlarging the convolution kernel size by skipping intermediate pixels and thus, requiring less $\texttt{multadds}$ operations.

The first work that suggested the use of learnable sized convolution kernels was box convolutions \cite{NIPS2018_7859}. The idea of using box filters with summed-area tables \cite{crow1984summed}, commonly known as integral images dates back many years and it is well-known to the Computer Vision community, as it became particularly popular with the work of \citet{Viola01robustreal-time} in object detection. The summed-area table can be efficiently parallelized using the Parallel Prefix Sum method \cite{10.1145/322217.322232}. This operation can be further accelerated as a hardware functional unit dedicated to compute the multi-parameter prefix-sum operation \cite{18021}. 

\citet{NIPS2018_7859} method is optimizing the kernel size parameters using approximate gradients by normalizing the sum by the area of the box. \citet{zhang2019accelerating} extended this idea by using interpolation to exploit non-integer coordinates. 
Inspired by this idea, we develop the proposed method for one-dimensional case of sequences. 
In contrast to the two previous methods, instead of using a fixed number of learnable sized kernels, we adaptively condition the size of the kernel on each input representation, effectively generating a different kernel size for each time-step token.

\begin{figure}[t]
\centering
\includegraphics[width=0.48\textwidth]{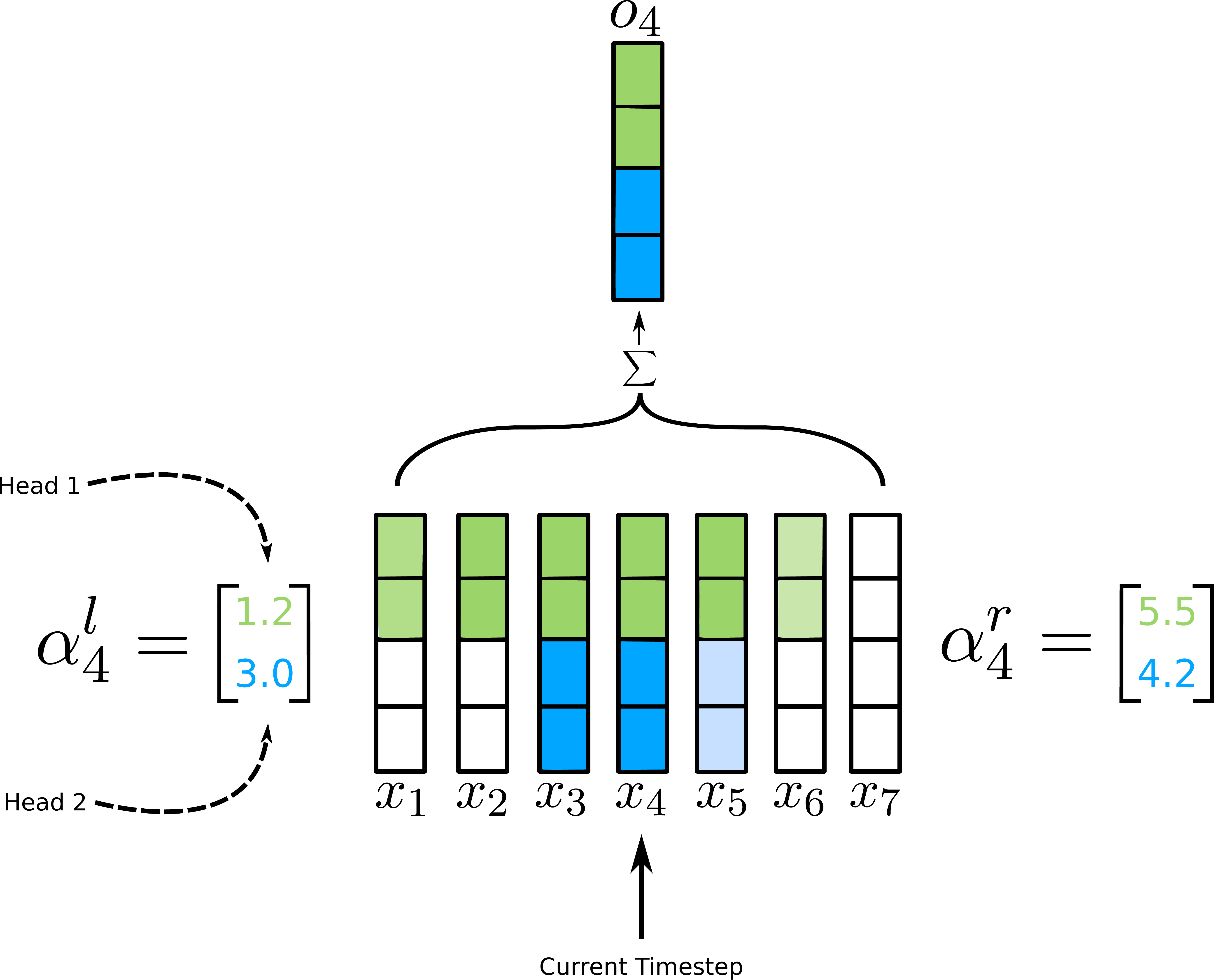}
\caption{The Time-aware Large Kernel convolution operation. For the current time-step, we compute the left and right offsets for each head, and then sum all the representation vectors inside these boundaries. This operation can be efficiently computed using summed-area tables with time complexity $O(\log(n))$ and compute the output representation for each time-step in $O(n)$ time.}
\label{fig:archs}
\end{figure}

\section{Methodology}
In this section, we present the proposed adaptive Time-aware Large Kernel (TaLK) Convolution method. First, we introduce the approach that computes a convolution operation using large kernels in $O(n)$ time, which assumes that left and right offsets are given. Next, we present our proposed method for generating offsets dynamically for each time-step. We then expand upon our method to use multiple heads and normalize the summed output vector. Next, we explain how to use TaLK Convolutions for decoding. Finally, we present the computational complexity analysis and comparison for the proposed method. Figure \ref{fig:archs} illustrates the {\talkfull} operation for a specific time-step during encoding.

\subsection{One-dimensional Large Kernel Convolution}
\label{sec:lkc}

Let $X = \{x_1, x_2, \ldots, x_n\}$ denote an input sequence, where $n$ is the length of the sequence, $x_i \in \mathbb{R}^d$ is the current input representation for the $i$-th word (i.e., the $i$-th time-step) and $d$ denotes the dimensionality of the vector representation (i.e., the number of channels).

The goal of this paper is to reduce the encoding time complexity for sequence modeling to $O(n)$. In other words, we set out to make the encoding at each time-step independent of the size of the receptive field. 
In addition, we want to explore alternative methods to the successful self-attention mechanism by equally using the number of neighbor tokens to represent a time-step instead of generating an attention distribution over the tokens. Specifically, we assume that simply summing the appropriate number of token representations is enough to represent the current time-step. That is, we encode the representation at the $i$-th time-step by
\begin{equation}
    o_{i} = \sum_{j=\alpha_{i}^l}^{\alpha_{i}^r}x_{j},
    \label{eq:naive_sum}
\end{equation}
where $1\leq\alpha_{i}^l\leq{i}\leq\alpha_{i}^r{\leq}n$ are the lower (left offset) and upper (right offset) bounds of the kernel size.

Applying Equation \ref{eq:naive_sum} for each time-step $i$ separately is inefficient since we do repetitive summations over the same representations. \citet{zhang2019accelerating} showed that using the summed-area table \cite{crow1984summed}, we can accelerate a summation convolution operation to any kernel size. Specifically, let $\mathcal{S} = \{\mathcal{S}_{0}, \mathcal{S}_{1}, \mathcal{S}_{2}, \ldots, \mathcal{S}_{n}\}$ be the summed-area table computed using
\begin{equation}
    \begin{cases} \mathcal{S}_{0} = 0, \\
    \mathcal{S}_{i} = \mathcal{S}_{i-1} + x_{i}, \quad 1\leq i\leq n. \end{cases}
\label{eq:sum_table}
\end{equation}

Given the left offset $\alpha_{i}^l$ and the right offset $\alpha_{i}^r$, we can compute the summation denoted as $o_i$ of the features between these offsets using the summed-area table
\begin{equation}
    o_{i} = \mathcal{S}_{a_{i}^{r}} - \mathcal{S}_{a_{i}^l-1}
    \label{eq:output}
\end{equation}

\subsection{Time-aware Large Kernel Generation}
\label{sec:talk}
Given the one-dimensional large kernel convolution above, it is important to determine the left and right offsets for computing representations at each time-step. The key of the proposed method is an adaptive time-aware large kernel convolution operation which has kernel sizes that vary over time as a learned function of the individual time steps; that is, we propose to learn the offsets of the summation kernel above for each time-step.

Specifically, we propose to use
a function $f^{\{l,r\}}$ : $\mathbb{R}^d \rightarrow \mathbb{R}$ 
to
generate for each $x_i$ the left $\tilde{a}_i^l$ and right $\tilde{a}_i^r$ relative offsets, 
where $\tilde{a}_i^{\{l,r\}} = \sigmoid(f^{\{l,r\}}(x_i)) \in [0,1]$. 
For each $\tilde{a}_i^{\{l,r\}}$ relative offset, we convert it to the absolute offset counterpart
in the following way
\begin{align}
\begin{split}
    a_{i}^l &= i - \tilde{a}_{i}^l \cdot l_\text{max} \\
    a_{i}^r &= i + \tilde{a}_{i}^r \cdot r_\text{max},
\end{split}
\label{eq:abs_off}
\end{align}
where $l_\text{max} \in \mathbb{Z}_{\geq 0}$ is the maximum allowed tokens to the left and $r_\text{max} \in \mathbb{Z}_{\geq 0}$ is the maximum allowed tokens to the right.

The absolute offsets up to this point represent real positive numbers. In the next step, we need to convert these numbers to integer indexes so we can select from the summed-area table using the Equation (\ref{eq:output}). Inspired by \citet{zhang2019accelerating}, we use one-dimensional interpolation to sample from the summed-area table 
by
using the  positive real-valued offsets $a_i^l, a_i^r$ as follows
\begin{align}
\begin{split}
    \mathcal{S}_{a_{i}^l-1} &= 
    \gamma^l \cdot \mathcal{S}_{\floor{a_{i}^l}-1} + 
    (1-\gamma^l) \cdot \mathcal{S}_{\ceil{a_{i}^l}-1}, \\
    \mathcal{S}_{a_{i}^r} &= 
    (1-\gamma^r) \cdot \mathcal{S}_{\floor{a_{i}^r}} + 
    \gamma^r  \cdot \mathcal{S}_{\ceil{a_{i}^r}},
\end{split}
\label{eq:off_use}
\end{align}
where $\floor{.}$ and $\ceil{.}$ are the floor and ceiling operators, $\gamma^l = \ceil{a_{i}^l} - a_{i}^l$ and $\gamma^r = a_{i}^r - \floor{a_{i}^r}$. The above equation is continuous and differentiable in the interpolation neighborhood. The partial derivatives of $\mathcal{S}_{a_{i}^{\{l,r\}}}$ with respect to $\tilde{a}_{i}^{\{l,r\}}$ are given by
\begin{align}
\begin{split}
    \frac{\partial \mathcal{S}_{a_{i}^l-1}}{\partial \tilde{a}_{i}^l} &=  l_\text{max} ( \mathcal{S}_{\floor{a_{i}^l}-1} - \mathcal{S}_{\ceil{a_{i}^l}-1}), \\
    \frac{\partial \mathcal{S}_{a_{i}^r}}{\partial \tilde{a}_{i}^r} &=  r_\text{max} ( \mathcal{S}_{\ceil{a_{i}^r}} - \mathcal{S}_{\floor{a_{i}^r}}).
\end{split}
\end{align}

The partial derivatives of $\mathcal{S}_{a_{i}^{\{l,r\}}}$ with respect to $\mathcal{S}_{\floor{a_{i}^{\{l,r\}}}}$ and $\mathcal{S}_{\ceil{a_{i}^{\{l,r\}}}}$ tokens are given by
\begin{equation}
\begin{aligned}
\frac{\partial \mathcal{S}_{a_{i}^l-1}}{\partial \mathcal{S}_{\floor{a_{i}^l}-1}} &= \gamma^l, \:&\:
\frac{\partial \mathcal{S}_{a_{i}^l-1}}{\partial \mathcal{S}_{\ceil{a_{i}^l}-1}} &= (1-\gamma^l), \\[5pt]
\frac{\partial \mathcal{S}_{a_{i}^r}}{\partial \mathcal{S}_{\floor{a_{i}^r}}} &= (1-\gamma^r), \:&\:
\frac{\partial \mathcal{S}_{a_{i}^r}}{\partial \mathcal{S}_{\ceil{a_{i}^r}}} &= \gamma^r.
\end{aligned}
\end{equation}

\begin{table*}[t]
\centering
\caption{
  Maximum path lengths, per-layer complexity and minimum number of sequential operations for different layer types. $n$ is the sequence length, $d$ is the representation dimension, $k$ is the kernel size of convolutions and $n_{buckets}$ is the number of hash buckets.}
\vskip 0.15in
\begin{tabular}{lccc}
\toprule
Layer Type & Complexity per Layer & Sequential & Maximum Path Length  \\
           &             & Operations &   \\
\hline
\rule{0pt}{2.0ex}Recurrent \cite{DBLP:journals/corr/SutskeverVL14} & $O(n \cdot d^2)$ & $O(n)$ & $O(n)$ \\
\makecell[l]{Convolutional\\ \cite{DBLP:journals/corr/KalchbrennerESO16,gehring2017convolutional}} & $O(k \cdot n \cdot d^2)$ & $O(1)$ & $O(\log_k(n))$ or $O(n/k)$ \\
Self-Attention \cite{vaswani2017attention} & $O(n^2 \cdot d)$ & $O(1)$ & $O(1)$ \\
Dynamic Convolutions \cite{wu2019pay} & $O(k \cdot n \cdot d)$ & $O(1)$ & $O(n/k)$ \\
Reformer \cite{kitaev2020reformer} & $O(n \cdot \log(n) \cdot d)$ & $O(\log(n))$ & $O(n/n_{buckets})$ \\
\midrule
{\talkconvfulls} (Ours) & $O(n \cdot d)$ & $O(\log(n))$ & $O(n/(l_\text{max} + r_\text{max} + 1))$ \\
\bottomrule
\end{tabular}
\label{tab:op_complexities}
\end{table*}

\subsection{Output Normalization and Offsets Dropout}
\label{sec:omod}
The idea of summing all the features in a window of size $[a_i^l, a_i^r]$ works well for shallow models. 
However, 
as the representation vectors at different time-steps are computed from summations over different numbers of neighbors,
their magnitudes of values can be different. 
As we introduce more layers, the disproportional magnitude of the inputs makes learning harder for the nodes in the layers that follow. 
To address this problem, we propose to 
normalize the output representations of {\talkconvfulls} 
as follows
\begin{equation}
    \tilde{o}_{i} = o_{i} \cdot \left(\frac{1}{l_\text{max}+r_\text{max}+1}\right).
\end{equation}
Such a simple window size based normalization can effectively get rid of the output magnitude differentiation problem 
resulted from summation kernels. 

In addition, we regularize the predicted offsets $\tilde{a}_i^{\{l,r\}}$ using Dropout \cite{DBLP:journals/corr/abs-1207-0580,JMLR:v15:srivastava14a}. Specifically, during training we drop out every predicted offset with probability $p$. This helps to prevent the model from quickly optimizing towards a specific window size and be able to generate more diverse offsets.

\subsection{Multi-headed Kernels}
\label{sec:mhk}

Although the offset computation above provides a mechanism that offers adaptive receptive fields for summation kernels at different time steps, a single pair of left and right offsets for all $d$ dimensions cannot yield good results, as different features might be related to their counterpart in the neighbor tokens in different way. Inspired by the idea of multi-head attention \cite{vaswani2017attention,wu2019pay}, we further propose to extend our proposed convolution kernel into a multi-head version by allowing different representation features, i.e., channels, to have different left and right offsets for each time-step. Moreover, instead of having entirely different convolution offsets across multiple channels, we adopt a depthwise version by separating the feature channels into multiple groups, each of which share the same pair of left and right offsets.

Specifically, we tie every subsequent number of $R=\frac{d}{H}$ channels together and group the channels into $H$ groups for each $x_i$, where $H$ is the number of heads. This results to $\hat{X} = \{\hat{x}_1, \hat{x}_2, \ldots, \hat{x}_n\}$, where $\hat{x}_i \in \mathbb{R}^{H{\times}R}$. Then we use a function $f^{\{l,r\}}$ : $\mathbb{R}^{H\times R} \rightarrow \mathbb{R}^{H}$  to generate for each $\hat{x}_i$ a vector of $H$ left relative offsets $\tilde{\bf \alpha}_i^l$ or right relative offsets $\tilde{\bf \alpha}_i^r$  via  $\tilde{\bf \alpha}_i^{\{l,r\}} = \sigmoid(f^{\{l,r\}}(\hat{x}_i)) \in [0,1]^H$.

\subsection{Decoding Using {\talkconvfulls}}
In an encoder/decoder sequence generation scheme \cite{DBLP:journals/corr/SutskeverVL14}, the encoder part of the model has access to both past and future tokens. The decoding part, however, must have access only to past tokens that are generated so far. Enforcing this with {\talkconvfulls} is straightforward by setting the $r_\text{max}$ value to zero.

\begin{figure}[t]
\centering
\includegraphics[width=0.4\textwidth]{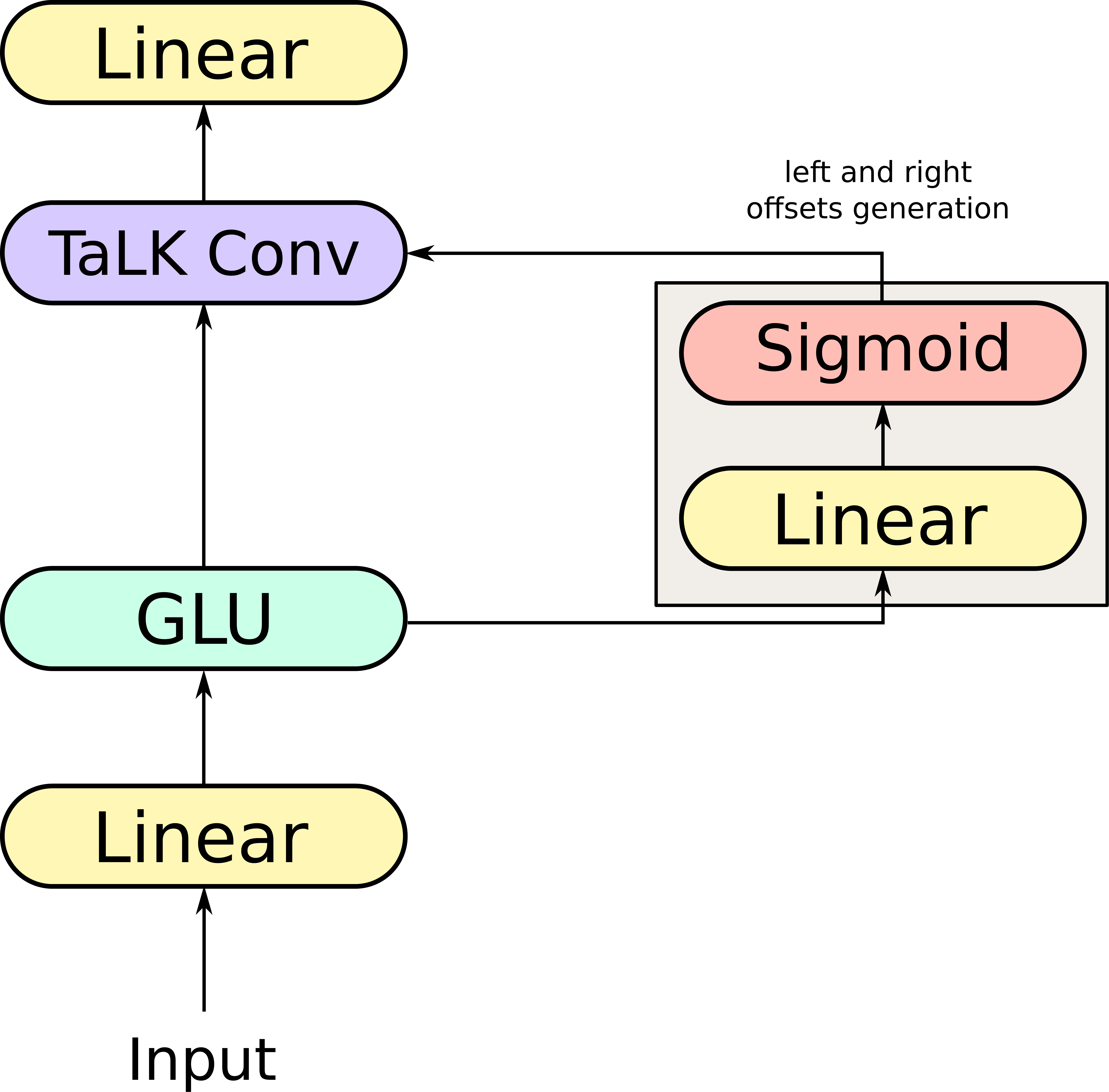}
\caption{The proposed \talkconvfull{} unit.}
\label{fig:unit_arch}
\end{figure}

\subsection{Module Architecture and Implementation}
\label{sec:arch}
For sequence modeling, we follow a similar module architecture as described in \citet{wu2019pay}. Specifically, we apply a linear layer to project the input embedding tokens from $d$ to $2d$ and then we apply a gated linear unit (GLU) \cite{DBLP:journals/corr/DauphinFAG16}. Next, we apply the \talkconvfull{} operation as described in Section \ref{sec:talk}. Finally, we apply a projection layer to the output representations from \talkconvfull{} with size $W \in \mathbb{R}^{d{\times}d}$. Figure \ref{fig:unit_arch} illustrates the \talkconvfull{} unit. In addition, we substitute all ReLU activation functions with the Swish function \cite{DBLP:journals/corr/abs-1710-05941}. Similar to \cite{vaswani2017attention,wu2019pay}, the overall architecture is composed by the TaLK Convolution unit which replaces the self-attention or the dynamic convolution unit followed by the position-wise feed-forward network (FFN) unit.

The summed-area table (Equation \ref{eq:sum_table}) can be efficiently computed on a GPU by performing a fast Parallel Prefix Sum \cite{10.1145/322217.322232} over the token dimension. This operation is usually efficiently implemented on modern deep learning frameworks under the name of cumulative sum. Applying the relative offsets to the summed-area table using core functions from deep learning frameworks is not a trivial task. Such an implementation is usually very inefficient leading to slower computation and memory overhead. For this reason, we implemented the operation using CUDA kernels that enabled us to parallelize the computation for each token.

\subsection{Computational Complexity}
\label{sec:tcc}
In this section we compare the complexity of 
the \talkconvfull{} operation against different modules for encoding an input sequence of representations. This comparison is shown on Table \ref{tab:op_complexities}. We follow a similar comparison as analyzed by \citet{vaswani2017attention}. Our comparison is based on three criteria: the time complexity of the operation, the amount of computations that can be executed in parallel and the path length between long-range dependencies. 

As shown in Table \ref{tab:op_complexities}, our proposed method requires the least number of operations. Specifically, it has a linear time complexity to encode a sequence and does not depend on hyper-parameter decisions such as the kernel size. In terms of the number of computations that can be parallelized, our method needs logarithmic time to compute the summed-area table (Equation \ref{eq:sum_table}). It is true that our method does not have a constant number of sequentially executed operations like all the other non-autoregressive counterpart methods, but the logarithmic time our method is requiring is inexpensive to compute even for very long sequences. 

It is shown by \citet{5264952} that a short path between any combination of token positions in the input and output sequences makes it easier to learn long-range dependencies. In practice, doubts have been cast over the ability of self-attention to model long-range dependencies \cite{DBLP:journals/corr/abs-1808-08946,wu2019pay}. \citet{wu2019pay} showed that using a limited context window can outperform self-attention. Our method has the advantage that the number of required computations is independent of the maximum window size and thus, it can be tuned or learned without extra cost. 

\begin{table*}[t]
\centering
\caption{Machine translation accuracy in terms of BLEU for \ende{} and \enfr{} on newstest2014.}
\vskip 0.15in
\begin{tabular}{lcrr}
\toprule
Model & Param (En-De) & \ende{} & \enfr{} \\
\midrule
\citet{gehring2017convolutional} & 216M & 25.2 & 40.5 \\
\citet{vaswani2017attention} & 213M & 28.4 & 41.0 \\
\citet{DBLP:journals/corr/abs-1711-02132} & 213M & 28.9 & 41.4 \\
\citet{DBLP:journals/corr/abs-1804-09849} & 379M & 28.5 & 41.0 \\
\citet{DBLP:journals/corr/abs-1803-02155}  & - & 29.2 & 41.5 \\
\citet{DBLP:journals/corr/abs-1806-00187} & 210M & 29.3 & \textbf{43.2} \\
\citet{wu2019pay} & 213M & \textbf{29.7} & \textbf{43.2} \\
\citet{kitaev2020reformer} & 213M & {29.1} & -- \\
\midrule
\talkconvfull{} (Ours) & 209M & 29.6 & \textbf{43.2} \\
\bottomrule
\end{tabular}
\label{tab:mt1}
\end{table*}

\begin{table}[t]
\centering
\caption{Machine translation accuracy in terms of BLEU on \iwslt{}.}
\vskip 0.15in
\begin{tabular}{lcr}
\toprule
Model & Param & \iwslt{} \\
\midrule
\citet{deng2018latent} & - & 33.1 \\
\citet{vaswani2017attention} & 47M & 34.4 \\
\citet{wu2019pay} & 43M & 35.2 \\
\midrule
\talkconvfull{} (Ours) & 42M & \textbf{35.5} \\
\bottomrule
\end{tabular}
\label{tab:mt2}
\end{table}

\section{Experiments}
\subsection{Datasets and Evaluation}
\label{sec:datasets}
We evaluated our proposed encoding technique on machine translation, abstractive summarization and language modeling. These three tasks are considered touchstone and challenging in the NLP field.

\paragraph{Machine Translation} On the machine translation task, we report results on three mainstream benchmark datasets: WMT English to German (En-De), WMT English to French (En-Fr) and IWSLT German to English (De-En). 

For all datasets, we replicated the pre-processing steps mentioned in \citet{wu2019pay}. Specifically, for the \ende{} we used the WMT'16 training data that consists of 4.5M sentence pairs. We validated on newstest2013 and tested on newstest2014. We employed byte-pair encoding (BPE) \cite{DBLP:journals/corr/SennrichHB15} to the sentences, with a 32K joint source and target vocabulary. For the \enfr{}, we used 36M training sentence pairs from WMT'14. We validated on newstest2012+2013 and tested on newstest2014 evaluation datasets. Using BPE, we generated a joint vocabulary between the source and the target languages of size 40K tokens. The \iwslt{} consists of 160K training sentence pairs. We lower cased all sentences and used a 10K joint BPE vocabulary.

For all datasets, we measured case-sensitive tokenized BLEU scores using $\texttt{multi-bleu}$\footnote{\url{https://github.com/moses-smt/mosesdecoder/blob/master/scripts/generic/multi\%2Dbleu.perl}}. Similarly to \citet{vaswani2017attention}, we applied compound splitting for \ende{}. We trained five random initializations of each model configuration and report test accuracy of the seed which resulted in the highest validation BLEU score. For all datasets, we used beam search with beam width 5. Similar to \citet{wu2019pay}, we tuned a length penalty as well as the number of checkpoints to average on the validation set.

\paragraph{Abstractive Summarization} For the abstractive summarization task, we decided to experiment with the \cnndm{} \cite{10.5555/2969239.2969428,nallapati-etal-2016-abstractive} dataset. The dataset is composed by approximately 280K news articles with associated multi-sentence summaries. We followed the same pre-processing steps as described by \citet{wu2019pay}. The BPE vocabulary consists of 30K subword tokens. We report results using the F1-Rouge (Rouge-1, Rouge-2 and Rouge-L) metric \cite{lin-2004-rouge}. For generating the summaries, similar to \citet{wu2019pay} we tuned the maximum output length, disallowing repeating the same trigram, and we apply a stepwise length penalty.

\paragraph{Language Modeling} We experimented on the \wiki{} \cite{DBLP:journals/corr/MerityXBS16} benchmark dataset. The training data contains approximately 100M tokens. A vocabulary of about 260K tokens was used, by discarding all tokens with a frequency below 3 as described in \citet{DBLP:journals/corr/MerityXBS16}. We followed \citet{DBLP:journals/corr/abs-1809-10853} and applied adaptive input representations. We replicated their setup and partition the training data into blocks of 512 contiguous tokens while ignoring document boundaries.

\begin{table*}[t]
\caption{Results on \cnndm{} abstractive summarization.}
\label{tab:abs}
\vskip 0.15in
\centering
\begin{tabular}{lcrrr}
\toprule
Model & Param & Rouge-1 & Rouge-2 & Rouge-L \\    
\midrule
LSTM \cite{DBLP:journals/corr/PaulusXS17} & - & 38.30 & 14.81 &  35.49  \\ 
CNN \cite{DBLP:journals/corr/abs-1711-05217} & - & 39.06 & 15.38 & 35.77 \\
Self-Attention Baseline \cite{wu2019pay} & 90M & 39.26 & 15.98 & 36.35  \\
Lightweight Convolution \cite{wu2019pay} & 86M & 39.52 & 15.97 & 36.51  \\
Dynamic Convolution \cite{wu2019pay} & 87M & 39.84 & 16.25 & 36.73 \\
\midrule
\talkconvfull{} (Standard) & 59M & 40.03 & 18.45 & 36.13 \\
\talkconvfull{} (Deep) & 83M & \textbf{40.59} & \textbf{18.97} & \textbf{36.81} \\ 
\bottomrule
\end{tabular}
\end{table*}

\begin{table}
\centering
\caption{Test perplexity on \wiki{}. We used adaptive inputs similar to \citet{DBLP:journals/corr/abs-1809-10853} and show that our method yields better perplexity than dynamic convolutions and comparative performance with self-attention.}
\vskip 0.15in
\begin{tabular}{lcrr}
\toprule
& Param & Test \\ \midrule
\citet{DBLP:journals/corr/GraveJU16} & - & 40.8 \\
\citet{DBLP:journals/corr/DauphinFAG16}  & 229M & 37.2 \\
\citet{DBLP:journals/corr/abs-1803-08240} & 151M & 33.0  \\
\citet{DBLP:journals/corr/abs-1803-10049} & - & 29.2 \\
\citet{DBLP:journals/corr/abs-1809-10853} & 247M & \textbf{20.5} \\
\midrule
Dynamic Convolutions & 255M & 25.0 \\
\talkconvfull{} (Ours) & 240M & 23.3 \\
\bottomrule
\end{tabular}
\label{tab:wiki_best}
\end{table}

\subsection{Experiment Details}
\paragraph{Hyper-Parameters} For the machine translation models, we followed the same hyper-parameter setup as described in \citet{wu2019pay}. Specifically, we follow for \ende{} and \enfr{} datasets the model hidden size $d$ was set to 1024, the feed-forward hidden size $d_{\text{ff}}$ was set to 4096 and the number of layers for the encoder and the decoder was set to 7 and 6 respectively. The number of heads was set to 16 and the $l_\text{max}, r_\text{max}$ values to $3,7,15,31{\times}4$ for each layer. For \iwslt{}, the model hidden size $d$ was set to 512, the feed-forward hidden size $d_{\text{ff}}$ was set to 1024 and the number of layers for the encoder and the decoder was set to 7 and 6 respectively. The number of heads was set to 4 and the $l_\text{max}, r_\text{max}$ values to $1,3,7,15{\times}4$ for each layer.

For the abstractive summarization models, we tested our method on two types of model configurations, the Standard and the Deep configurations. Both settings have a hidden size $d$ of 512, a feed-forward hidden size $d_{\text{ff}}$ of 2048 and number of heads to be 8. The Standard model has 7 encoder and 6 decoder layers with the $l_\text{max}, r_\text{max}$ values be $3,7,15,31{\times}4$ for each layer. The Deep model has 10 layers for both the encoder and decoder with the $l_\text{max}$ values be $3,7,15,31{\times}7$ and the $r_\text{max}$ be $3,7,31{\times}8$.

For the language model, we followed the same configuration as \citet{DBLP:journals/corr/abs-1809-10853}. We used 17 decoding layers, each layer with a 1024 hidden size, a 4096 feed-forward hidden size and 8 heads. The adaptive input factor was set to 4. The $l_\text{max}$ values were set to $3,7,15,31,63{\times}12$ and $r_\text{max}$ to zero for each layer.

\begin{table*}
\centering
\caption{Throughput and memory consumption decrease measured for different sequence lengths ($n$) on a batch of size 10 with each token being represented with $d=1024$ and $H=16$. Throughput is calculated across 100K iterations of a single input encoding execution for each method. Memory decrease is computed as how many times less memory we need to encoding the input embedding compared to Self-Attention. Larger numbers indicate better performance.}
\vskip 0.15in
\begin{tabular}{l|cc|cc|cc|cc}
\toprule
\multirow{2}{*}{Method} & \multicolumn{2}{c|}{$n=10$} & \multicolumn{2}{c|}{$n=100$} & \multicolumn{2}{c|}{$n=1,000$} & \multicolumn{2}{c}{$n=10,000$} \\
\cline{2-9}
& iter/sec & Mem. $\downarrow$ & iter/sec  & Mem. $\downarrow$ & iter/sec  & Mem. $\downarrow$ & iter/sec & Mem. $\downarrow$ \\ \midrule
Self-Attention & 4576 & 1x & 3437 & 1x & 102 & 1x & OOM & 1x \\
DynamicConv ($k=3$) & 3739 & 1x & 3308 & 0.99x & 443 & 2.8x & 45 & 25.4x \\
DynamicConv ($k=31$) & 4535 & 0.97x & 3860 & 1x & 325 & 2.7x & 29 & 20.2x \\
\midrule
\talkconvfull{} & \textbf{9686} & \textbf{1.1x} & \textbf{6126} & \textbf{1.1x} & \textbf{898} & \textbf{3.1x} & \textbf{92} & \textbf{26.4x} \\
\bottomrule
\end{tabular}
\label{tab:time_mem_table}
\end{table*}

\paragraph{Optimization} We used the Adam optimizer \cite{kingma2014adam} with default values. In addition, our models were optimized using the cosine learning rate schedule \cite{loshchilov2016sgdr}. We linearly warmed up for 10K steps from $10^{-7}$ to $10^{-3}$. For \iwslt{}, we used the inverse square root learning rate schedule \cite{vaswani2017attention}. We set the dropout to 0.3 for \ende{} and \iwslt{} and 0.1 for \enfr{}.

For the \ende{} and \enfr{} benchmarks, the batch size was set to 3,584 tokens per batch, per GPU. We accumulated the gradients for 16 batches before applying an update which results in an effective batch size of 450K tokens. We trained the \ende{} model for 30K steps and the \enfr{} for 80K steps. For \iwslt{}, we trained on a single GPU with 4,000 maximum tokens per batch  for 50K steps. For the abstractive summarization models we followed the same setup as in \citet{wu2019pay} and for the language model the same setup as in \citet{DBLP:journals/corr/abs-1809-10853}.
    
\paragraph{Hardware Details} We trained the \ende{}, \enfr{}, \cnndm{} and \wiki{} models on 8 NVIDIA RTX 2080 Ti GPUs using mixed-precision training \cite{DBLP:journals/corr/abs-1710-03740} and the \iwslt{} model using a single GPU. We employed our own CUDA implementation, wrapped as a standalone PyTorch layer for the \talkconvfull{} operation. All experiments were run using the Fairseq \cite{ott2019fairseq} toolkit.

\subsection{Results on Machine Translation}

We demonstrate the effectiveness of our model in the \ende{} and \enfr{} translation benchmarks. Table \ref{tab:mt1} shows that our method is able to achieve comparable results to current state-of-the-art methods. Specifically, our method was able to match the state-of-the-art score on \enfr{}, a benchmark dataset that is considered indicative for the effectiveness of a method due to the large number of training examples (36M) it contains. Additionally for \ende{}, our method is only 0.1 BLEU points behind the current state-of-the-art score. 
It is important to underline, however, that our method uses the least number of parameters compared to the other counterpart methods.

Table \ref{tab:mt2} shows results for \iwslt{} benchmark dataset. Following \citet{wu2019pay}, we employed a smaller model with less parameters to reflect the size of the dataset. Specifically, we set $d$ to 512, $d_{\text{ff}}$ to 1024 and $H$ to 4. Furthermore, we disabled the GLU unit that is described in Section \ref{sec:arch} and made the input projection layer to size $W \in \mathbb{R}^{d{\times}d}$. Our method was able to outperform all other methods setting a new state-of-the-art result. 

\subsection{Results on Abstractive Summarization}
We evaluated the proposed method on the task of abstractive summarization. We test the method's ability to process long documents on the \cnndm{} dataset. We encode an article of up to 400 sub-words and we generate a summarization composed from multiple sentences. Table \ref{tab:abs} shows the results of our experiments. Our Standard model is able to achieve better results on the Rouge-1 and Rouge-2 metrics than previous methods. In addition, the Standard model is using significantly less parameters, approximately 30M parameters less. The Deep model uses more layers to closely match the number of parameters and is able to outperform all previous models. This shows that our method is able to encode long sequences successfully without having the need to have access to all context.

\subsection{Results on Language Modeling}
We evaluated our method on the task of language modeling. We considered the \wiki{} benchmark dataset and we compared against recent methods in the literature. Particularly, we followed the setup that was implemented in the adaptive inputs baseline \cite{DBLP:journals/corr/abs-1809-10853}. This work suggest the use of self-attention with adaptive input representations. We substituted the self-attention module with our method. In order to assimilate the number of parameters used in their experiments, we increased the number of layers by one. As seen on Table \ref{tab:wiki_best}, our method yields better perplexity than dynamic convolutions when trained using the same settings, including the same maximum kernel size. In addition, we get comparative performance with self-attention. Moreover, we use less number of parameters than the best comparison methods.

\begin{table*}[t]
\centering
\caption{Ablation on \iwslt{} validation set. (+) indicates that a result includes all preceding features.}
\vskip 0.15in
\begin{tabular}{lcr}
\toprule
Model & Param & BLEU \\
\midrule
\talkconvfull{} ($a_i^l,a_i^r$=1x7, $H$=1) & 42M & diverges \\
+ Output Normalization & 42M & 35.70 $\pm$ 0.1 \\
+ Increasing Max Offsets ($a_i^l,a_i^r$=1,3,7,15x4) & 42M & 36.23 $\pm$ 0.1 \\
+ Offsets Dropout ($p$=0.1) & 42M & 36.37 $\pm$ 0.05 \\
+ Fully-headed Kernels ($H$=512) & 47M & 36.51 $\pm$ 0.07 \\
+ Multi-headed Kernels ($H$=4) & 42M & \textbf{36.65} $\pm$ 0.05 \\
\midrule
Replacing Swish with ReLU & 42M & 36.21 $\pm$ 0.05 \\
\bottomrule
\end{tabular}
\label{tab:ablation}
\end{table*}

\subsection{Encoding Inference Speed Comparison}
We also compared our method against other non-autoregressive methods in terms of encoding inference speed and memory consumption. We measured the speed using a single NVIDIA RTX 2080 Ti GPU with full precision floating-point arithmetic (FP32). Specifically, we measured the throughput of encoding a batch of size $B=10$, $d=1024$ and $H=16$. For each method, we only took into consideration the time it takes to process using the core approach of each encoding method. 

For self-attention \cite{vaswani2017attention}, we only timed the attention operation $\mathrm{softmax}(\frac{QK^T}{\sqrt{d_k}})V$. For dynamic convolutions \cite{wu2019pay}, we only timed the operation $\text{DepthwiseConv}(X, \softmax(\emW_\text{dyn}), i, c)$ where $W_\text{dyn} \in \mathbb{R}^{n{\cdot}B{\cdot}H{\times}K}$ is the generated kernel for each time-step. The authors of dynamic convolutions proposed two ways of implementing their method. The first method uses the standard convolution unfolding function which is faster for longer sequences. The second approach is the band matrix trick method which copies and expands the normalized weights matrix into a band matrix. This second approach yields faster execution time for shorter sequences but is more memory intensive. In order to be fair, in our experiments we used unfolding to sequences longer than 500 tokens and band matrices for shorter sequences. We also set $K$ to 3 and 31, the first being the smallest kernel size dynamic convolutions use and the second being the largest. Finally, for our method we measured the time to compute the large kernel convolution operation given the relative offsets. 
We evaluated for 100K iterations across four different sequence lengths $n$.

Table \ref{tab:time_mem_table} shows that our method yields much better throughput than all other methods. Specifically, the number of iterations of self-attention per second is comparable to dynamic convolutions for short sentences ($n<500$). Our method allows for more sentences to be processed each second, leading to a much higher throughput. For longer sentences, self-attention is notably slower than our method and for the case of $n=10,000$, self-attention was running out-of-memory and was not able to execute an iteration. Although our method has a logarithmic time complexity for computing the summed-area table (Section \ref{sec:tcc}), the fact that we are computing a much more inexpensive in terms of complexity operation, specifically we only utilize additions, other methods with $O(1)$ complexity are less efficient due to the employment of both multiplication as well as addition operations. Therefore, our method has a considerably higher throughput compared to dynamic convolutions. 

Furthermore, we examined the running memory requirements for all three different non-autoregressive methods. We compared dynamic convolutions and our proposed method against self-attention and report the number of times we reduced the running memory compared to self-attention. For all sequence length cases, our method requires less memory than dynamic convolutions when compared to the ``expensive'' self-attention operation. The times we were able to decrease the memory consumption can be seen on Table \ref{tab:time_mem_table}.

\subsection{Model Ablation}

In order to evaluate the importance of the different choices for the \talkconvfulls{}, we varied our baseline model, described in Section \ref{sec:talk}, using the different proposed extensions mentioned in Sections \ref{sec:omod} and \ref{sec:mhk}. We measured the performance on the validation set of the \iwslt{} translation benchmark dataset. We used beam search as described in Section \ref{sec:datasets}. We report the results in Table \ref{tab:ablation}.

Initially, we modified the baseline model with the addition of the output normalization (Section \ref{sec:omod}). As seen in Table \ref{tab:ablation}, the original method is not able to converge. This validates our intuition that since we are summing the available information inside the kernel, not normalized outputs make learning difficult for the layers that follow. Next, we increased the values $l_\text{max},r_\text{max}$ to allow larger adaptive kernel sizes which yielded a higher performance without additional computation cost. Further, we introduced a dropout unit with probability $p=0.1$ on the generated relative offsets. This allowed for the performance to increase further as we stopped the model from overfitting over the same window size. Next, we increased the number of heads $H$ from 1 to 512 (all available dimensions) and we called this fully-head \talkconvfull{}. We can see that by treating each of the 512 dimensions separately and generating 512 relative offsets, we were able to increase the performance. However, we believe that by having each dimension generate its own offsets actually brings some noise. Thus, we reduced the number of heads to $H=4$ which increased the performance even more. Finally, we show that by substituting the Swish activation function with the ReLU function the performance drops which justifies our decision to use the former.

\section{Conclusion}
In this work, we presented \talkfullsabbr{}, a novel adaptive convolution method based on summation kernel for sequence representation and encoding. It learns to predict the kernel boundaries for each time-step of the sequence. In contrast to all other non-autoregressive methods, this approach needs true linear time $O(n)$ with respect to the sequence length, while being able to successfully encode a sequence without using the notion of attention. We validated the proposed method on three NLP tasks, machine translation, abstractive summarization and language modeling, and achieved a comparative performance. Moreover, we showed both analytically and empirically that the proposed method is faster than previous approaches and that it is able to encode longer sentences quicker and with a smaller running memory footprint. For future work, we plan to apply our method to other sequential tasks such as question answering and the PG-19 benchmark dataset \cite{Rae2020Compressive}. We will also explore this novel convolution mechanism in the area of computer vision.

\section*{Acknowledgements}
This research was supported by the Canada Research Chairs program and the NSERC Discovery grant. We would like to express our gratitude to our anonymous reviewers for their valuable comments and feedback. A special thank you to Vasileia Karasavva for editing and proofreading the final manuscript.

\bibliography{example_paper}
\bibliographystyle{icml2020}

\end{document}